\documentclass[10pt,twocolumn,letterpaper]{article}

\usepackage{iccv}
\usepackage{times}
\usepackage{epsfig}
\usepackage{graphicx}
\usepackage{amsmath}
\usepackage{amssymb}
\usepackage{array}
\usepackage{pifont}
\usepackage{capt-of,etoolbox}
\usepackage{multirow}
\usepackage{cite}
\usepackage{hhline}
\usepackage{authblk}
\usepackage{bbding}
\usepackage{algorithm}
\usepackage{algorithmic}
\usepackage[utf8x]{inputenc} 
\usepackage{braket}
\usepackage{boldline}
\newcommand{\cmark}{\ding{51}}%
\newcommand{\xmark}{\ding{55}}%
\usepackage[table,xcdraw]{xcolor}
\usepackage[pagebackref=true,breaklinks=true,letterpaper=true,colorlinks,bookmarks=false]{hyperref}

\iccvfinalcopy % *** Uncomment this line for the final submission

\def\iccvPaperID{2657}

\begin{document}

%%%%%%%%% TITLE
% \title{DistDepth: Structure Distillation-Guided Self-Supervised Monocular Indoor Depth Estimation}
\title{Learning Depth from Habitat: Self-Supervised Monocular Indoor Depth}
\title{How to Learn Depth in a Self-Supervised Fashion for Indoor?}
\title{Structured and Accurate Self-Supervised Indoor Depth}
\title{Closing Sim-to-Real Gap for Self-Supervised Indoor Depth}
\title{Heterogeneous Self-Supervised Monocular Indoor Depth Estimation}
\title{DistDepth: Structure Distillation-Guided Self-Supervised Monocular Indoor Depth Estimation}
\title{Structured and Metric-Accurate Self-Supervised Monocular Indoor Depth}
% \title{Generalizability Enhancement for Self-Supervised Monocular Indoor Depth}
\title{Learning Depth from Habitat: Self-Supervised Monocular Indoor Depth}
\title{Toward Practical Monocular Indoor Depth Estimation}

\author{Cho-Ying Wu$^{2}$, Jialiang Wang$^{1}$, Michael Hall$^{1}$, Ulrich Neumann$^{2}$ and Shuochen Su$^{1}$ \\
$^1$Meta Reality Labs, $^2$University of Southern California\\
{\tt\small \{jialiangw,michaelhall,shuochsu\}@fb.com, }
{\tt\small \{choyingw, uneumann\}@usc.edu}
}
\makeatletter
\let\@oldmaketitle\@maketitle
\renewcommand{\@maketitle}{\@oldmaketitle
  \centering\includegraphics[width=0.95\linewidth]{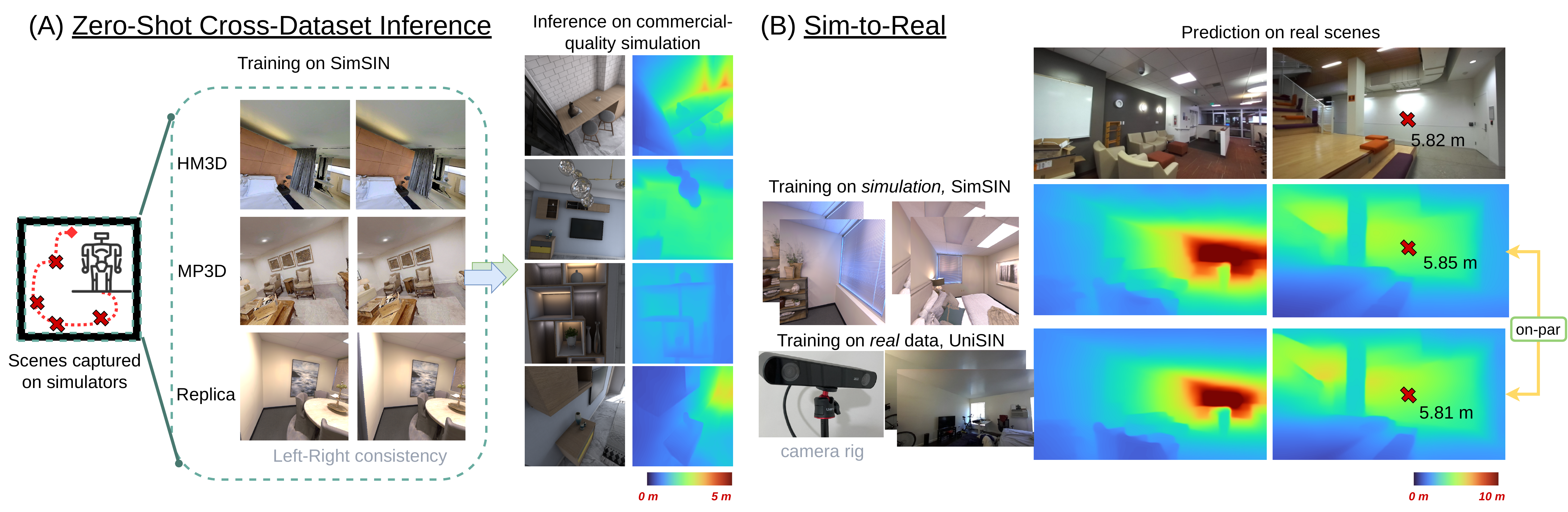}
  \captionof{figure}{\textbf{Advantages of our framework.} (A) We attain zero-shot cross-dataset inference. (B) Our framework trained on simulation data produces on-par results with the one trained on real data.\\}
  \label{teaser}}

\maketitle
\begin{abstract}
The majority of prior monocular depth estimation methods without groundtruth depth guidance focus on driving scenarios. We show that such methods generalize poorly to unseen complex indoor scenes, where objects are cluttered and arbitrarily arranged in the near field. To obtain more robustness, we propose a structure distillation approach to learn knacks from an off-the-shelf relative depth estimator that produces structured but metric-agnostic depth. By combining structure distillation with a branch that learns metrics from left-right consistency, we attain structured and metric depth for generic indoor scenes and make inferences in real-time. To facilitate learning and evaluation, we collect SimSIN, a dataset from simulation with thousands of environments, and UniSIN, a dataset that contains about 500 real scan sequences of generic indoor environments. We experiment in both sim-to-real and real-to-real settings, and show improvements, as well as in downstream applications using our depth maps. This work provides a full study, covering methods, data, and applications aspects.
\end{abstract}

\section{Introduction}
\label{intro}
This work proposes a \textbf{practical} indoor depth estimation framework that has the following features: \textit{learning from off-the-shelf estimators and left-right image pairs without their depth annotations}, \textit{efficient training data collection}, \textit{high generalizability to cross-dataset inference}, and \textit{accurate and real-time depth sensing}. 
Our work applies to consumer-level AR/VR, such as 3D indoor scene reconstruction and virtual object insertion and interaction with environment ~\cite{Occlusion2018}

Although self-supervised depth estimation, especially using left-right consistency, has attracted much research interest recently, popular works, such as MonoDepth \cite{godard2017unsupervised}, MonoDepth2 \cite{Godard_2019_ICCV}, DepthHints \cite{watson2019self}, and ManyDepth \cite{watson2021temporal}, mainly focus on driving scenes and are trained on large-scale driving datasets like KITTI \cite{geiger2012we} and Cityscapes \cite{Cordts2016Cityscapes}, and it is unclear how these methods apply on indoor environments. Learning \textit{indoor} depth via self-supervision is arguably more challenging for a number of reasons: (1) \textit{structure priors}: depth estimation for driving scenes imposes a strong scene structure prior to the learning paradigm. The upper parts of images, commonly occupied by the sky or buildings, are typically farther away; on the other hand, the lower parts are usually roads extending to the distance \cite{Dijk_2019_ICCV}. By contrast, the structure priors are much weaker for indoor environments since objects can be cluttered and arranged arbitrarily in the near field. (2) \textit{distribution}: scene depth for driving scenarios tends to distribute more evenly across near to far ranges on roads, whereas indoor depth can be concentrated in either near or far ranges, such as zoom-in views of desks or ceilings. The uneven depth distribution makes it challenging to predict accurate metric depth for indoor scenes. (3) \textit{camera pose}: depth-sensing devices can move in 6DoF for indoor captures, but they are typically anchored on cars for collecting driving data where translations are usually without elevation and rotations are dominated by yaw angle. Therefore, a desirable network needs to be more robust to arbitrary camera poses and complex scene structures for indoor cases. (4) \textit{untextured surfaces}: large untextured regions, such as walls, make the commonly used photometric loss ambiguous.

In this work we propose \textit{DistDepth}, a structure distillation approach to enhance depth accuracy trained by self-supervised learning. DistDepth uses an off-the-shelf relative depth estimator, DPT \cite{Ranftl2020, Ranftl2021} that produces \textit{structured but only relative depth} (output values reflect depth-ordering relations but are metric-agnostic). Our structure distillation strategy encourages depth structural similarity both statistically and spatially. In this way,  depth-ordering relations from DPT can be effectively blended into metric depth estimation branch trained by left-right consistency. Our learning paradigm only needs an off-the-shelf relative depth estimator and stereo image inputs without their curated depth annotations. Given a monocular image at test time, our depth estimator can predict structured and metric-accurate depth with high generalizability to unseen indoor scenes (Sec.~\ref{sec:distdepth}). Distillation also helps downsize DPT's large vision transformer to a smaller architecture, which enables real-time inference on portable devices.

We next describe our dataset-level contributions. Current publicly available stereo datasets are either targeting driving scenarios~\cite{geiger2012we, Gaidon:Virtual:CVPR2016, Cordts2016Cityscapes, Argoverse, wilson2021argoverse}, small-scale and lacking scene variability~\cite{scharstein2014high, schops2019bad}, rendered from unrealistic-scale 3D animations~\cite{mayer2016large, Butler2012}, or collected in-the-wild~\cite{wang2019web, hua2020holopix50k}.
Popular indoor datasets are either small-scale (Middlebury \cite{scharstein2014high}) or lacking stereo pairs (NYUv2 \cite{SilbermanECCV12}). There is currently no large-scale indoor stereo dataset to facilitate left-right consistency for self-supervised studies. We utilize the popular Habitat simulator \cite{savva2019habitat, szot2021habitat} to collect stereo pairs in 3D indoor environments. Commonly-used environments are chosen, including Replica \cite{straub2019replica}, Matterport3D (MP3D) \cite{chang2017matterport3d}, and Habitat-Matterport 3D (HM3D) \cite{ramakrishnan2021habitat}, to create \textit{SimSIN}, a novel dataset consisting of about 500K simulated stereo indoor images across about 1K indoor environments (Sec.~\ref{sec:dataset}). With SimSIN, we are able to investigate performances of prior self-supervised frameworks on indoor scenes \cite{Godard_2019_ICCV, watson2019self, watson2021temporal}. We show that we can fit on SimSIN by directly training those models, but such models generalize poorly to heterogeneous domain of unseen environments. Using our structure distillation strategy, however, can produce highly structured and metric-accurate depth on unseen data (Sec.~\ref{sec:analysis}).

Several commercial-quality simulations and real data are utilized for evaluation, including a challenging virtual apartment (VA) sequence \cite{UE4Environment, UnrealEngine4}, pre-rendered scenes in Hypersim \cite{roberts2020hypersim}, and real monocular images in NYUv2 \cite{silberman2012indoor}. To further investigate the gap between training on simulation v.s. training on real data, we further collect \textit{UniSIN}, a dataset including 500 real stereo indoor sequences, amounting to 200K images, in a university across buildings and spaces using off-the-shelf high-performing stereo cameras. We show that our DistDepth trained on simulation data only has on-par performance with those trained on real data. 

Our DistDepth is especially capable of \textit{1. attaining zero-shot cross-dataset inference}, and \textit{2. closing the gap between sim-to-real and real-to-real learning}, as shown in Fig.~\ref{teaser}. Throughout the work we visualize depth maps in actual \textit{metric} ranges unless marked as relative depth.
We summarize our contributions as follows.

1. We propose DistDepth, a framework that distills depth-domain structure knowledge into a self-supervised depth estimator to obtain highly structured and metric-accurate depth maps. 

2. We present SimSIN, a large-scale indoor simulation dataset that fuels the study of indoor depth estimation via left-right consistency, and a real dataset, UniSIN, that targets at studying the gap between training on simulation and real data.

3. We attain a practical indoor depth estimator: learning without curated depth groundtruth, efficient and effective data collection by simulation, high generalizability, and accurate and real-time inference for depth sensing. 

\section{Related Work}
\label{sec:related}
\textbf{Monocular Scene Depth Estimation}. Much research interest focuses on learning-based methods to learn a mapping: $\mathcal{I}\to \mathcal{D}$, from image to depth domain. 

\begin{figure*}[hbt]
    \centering
    \includegraphics[width=0.95\linewidth]{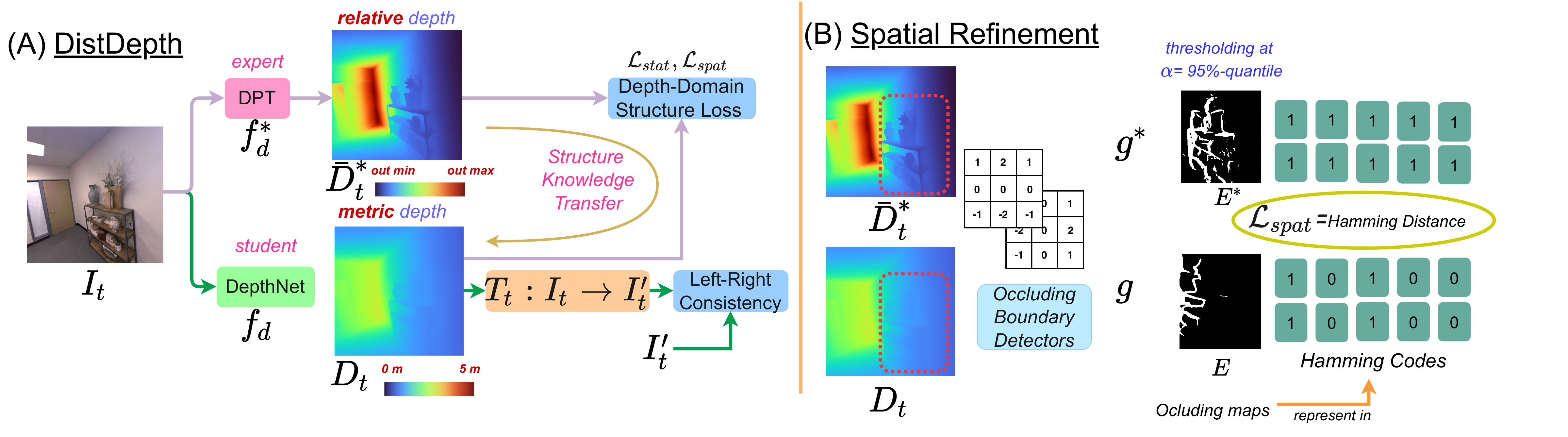}
    \vspace{-8pt}
    \caption{\textbf{DistDepth overview.} We distill structures from an off-the-shelf expert to a self-supervised depth estimation branch, DepthNet. Such an approach enables us to obtain metric depth maps with fine structures and still work without curated depth annotations. Note that we omit the temporal warping and PoseNet here for simplicity.}
    \vspace{-13pt}
    \label{distdepth}
\end{figure*}

\subsection{Supervised Scene Depth Estimation}
\label{sec:supervised}
Supervised learning requires pixel-level depth annotation. Early methods~\cite{eigen2014depth, xu2018structured, eigen2015predicting, liu2015deep, laina2016deeper, fu2018deep,yin2019enforcing,lee2019big,ramamonjisoa2019sharpnet,Hu2019RevisitingSI} perform pixel-level regression of depth values using convolutional neural networks to minimize a loss between predictions and groundtruth. Recently, Bhat \etal~\cite{bhat2021adabins} adopt adaptive bins for depth regression and use vision transformers \cite{dosovitskiy2020image}. MiDaS \cite{Ranftl2020} and Wei \etal~\cite{yin2021learning} smartly mix several datasets to attain large-scale depth training using scale-and-shift-invariant losses. BoostingDepth \cite{miangoleh2021boosting} fuses multi-scale depth cues in MiDaS based on observations from \cite{hu2019visualization}, but it takes minutes to post-process a depth map. DPT (MiDaS-V3) \cite{Ranftl2021} designs a dense vision transformer and achieves better results than the original MiDaS. 

Although state-of-the-art MiDaS \cite{Ranftl2020} and DPT~\cite{Ranftl2021} can estimate fine-grained depth structures for \textit{in-the-wild} images, they only provide relative depth, which is up to unknown scale and shift factors to align with actual size due to the mixed-dataset training strategy. Our DistDepth adopts such pretraining on in-the-wild scenes as an expert for depth structure distillation to attain both structured and metric depth from a branch trained by left-right consistency.

\subsection{Left-Right and Temporal Consistency}
\label{sec:self-supervised}
Left-Right and temporal consistency help attain self-supervised learning to lift requirements on groundtruth depth for training, which steps closer to practical depth sensing \cite{klingner2020self, guizilini20203d, kumar2020unrectdepthnet, Kumar_2021_WACV, poggi2020uncertainty}. MonoDepth \cite{godard2017unsupervised} learns depth from stereo pairs and uses left-right depth reprojection with photometric loss minimization. MonoDepth2 \cite{Godard_2019_ICCV} further includes temporal neighboring frames and also minimizes photometric consistency losses. DepthHints \cite{watson2019self} adopts pre-computed SGM \cite{hirschmuller2006stereo, hirschmuller2007stereo} depth from stereo pairs as a proxy and still stay self-supervised. ManyDepth \cite{watson2021temporal} uses test-time multi-frame inputs with cost-volume minimization to obtain more accurate predictions. However, these methods all focus on driving scenarios, and their applicability to indoor data is yet to be investigated. Our work is based on the left-right and temporal consistency with depth structure distillation to attain structured, metric, and generalizabile depth estimation, while most works do not discuss generalizability \cite{yang2020d3vo, gu2021dro, gordon2019depth, Godard_2019_ICCV,watson2021temporal}. Note that our structure distillation is different from regular distillation \cite{hinton2015distilling,zagoruyko2016paying, romero2014fitnets} since our expert is capable of only estimating depth ordering, and it needs to be combined with metrics inferred by the student to obtain the end output.
Some other works attain self-supervision by only temporal consistency \cite{bian2021unsupervised,  ji2021monoindoor}, which makes scale less robust. Another work \cite{li2021structdepth}, based on Manhattan world assumption, minimizes co-planar and normal losses but only shows robustness to planar regions with inherently ambiguous scale. 

\section{Method}
\label{sec:method}
\subsection{Basic Problem Setup}
\label{sec:setup}
We describe the commonly adopted left-right and temporal photometric consistency in self-supervised methods such as MonoDepth2, DepthHints, and ManyDepth in this section.
During training, $I_t$ and $I'_t$ are stereo pairs at timestep $t$. DepthNet $f_d$ is used to predict depth of $I_t$, $D_t = f_d(I_t)$. With known camera intrinsic $K$ and transformation $T_t: I_t \to I'_t$ using stereo baseline, one can back-project $I_t$ into 3D space and then re-project to the imaging plane of $I'_t$ by utilizing $K$, $D_t$, and $T_t$. $\hat{I}'_t = I_t \braket{proj(D_t, T_t, K)}$ denotes the reprojection. The objective is to minimize photometric loss $\mathcal{L}= pe(I'_t, \hat{I}'_t)$, where $pe$ is shown as follows.
\begin{equation}
    \vspace{-4pt}
     pe(I'_t, \hat{I}'_t) = \kappa \frac{1-\text{SSIM}(I'_t, \hat{I}'_t)}{2} + (1-\kappa)L_1(I'_t, \hat{I}'_t),
     \vspace{-3pt}
\label{photometric_loss}
\end{equation}
where $\kappa$ is commonly set to 0.85, SSIM \cite{wang2004image} is used to measure the image-domain structure similarity, $L_1$ is used to compute the pixel-wise difference. $pe(I'_t, \hat{I}'_t)$ measures photometric reconstruction error of a stereo pair to attain left-right consistency.

Temporal neighboring frames are also utilized to compute photometric consistency. PoseNet calculates relative camera pose between timestep $t$ and $t+k$: $T_{t+k\to t}=f_p(I_t, I_{t+k})$ with $k \in \{1,-1\}$. Then, temporal consistency is attained by warping an image from $t+k$ to $t$ and calculating photometric consistency in Eq.~\ref{photometric_loss}. At inference time, depth is predicted from a monocular image via $D=f_d({I})$.

\begin{figure}[bt!]
    \centering
    \includegraphics[width=1.0\linewidth]{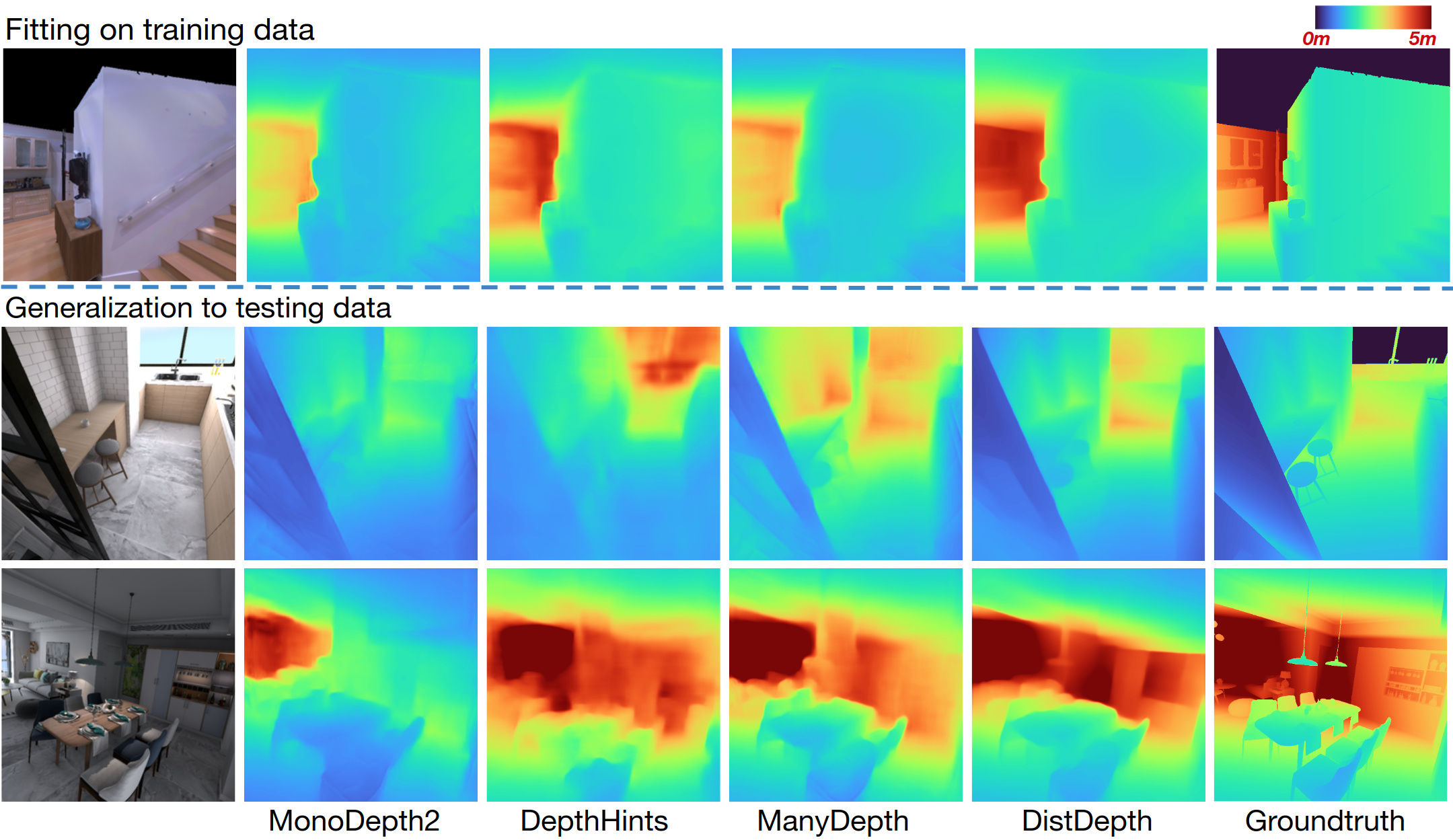}
    \vspace{-16pt}
    \caption{\textbf{Intra-/ Inter-Dataset inference.} Prior self-supervised works can fit training data (SimSIN) shown in the first row, but they generalize poorly to unseen testing dataset (VA), shown in the second and third rows. Our DistDepth can produce more structured and accurate ranges with reference to groundtruth.}
    \vspace{-11pt}
    \label{prior_work}
\end{figure}

\textbf{Applicability}. We train MonoDepth2, DepthHints, and ManyDepth on the SimSIN dataset and exemplify the scene fitting later in Fig.~\ref{prior_work}. Prior arts fit the training set but do not generalize well for cross-dataset inference due to unseen complex object arrangements for indoor environments.

\subsection{DistDepth: Structure Distillation from Expert}
\label{sec:distdepth}
To overcome the generalizability issue when applying self-supervised frameworks to indoor environments, we propose DistDepth (Fig.~\ref{distdepth}). DPT \cite{Ranftl2021} using dense vision transformer can produce highly structured but only relative depth values by $D_t^*=f_d^*(I_t)$\footnote{DPT (and MiDaS) outputs relative relation in disparity (inverse depth) space since it trains on diverse data sources (laser-based depth, depth from SfM, or stereo with unknown calibration). We inverse its outputs and compute losses in the depth space since our training data source is single} as explained in Sec.~\ref{sec:supervised}. We extract the depth-domain structure of $D_t^*$ and transfer to the self-supervised learning branches, including DepthNet $f_d$ and PoseNet $f_p$. The self-supervised branch learns metric depth since it leverages stereo pairs with known camera intrinsic and baseline with depth warping operation $I_t \braket{proj(D_t, T_t, K)}$. Our distillation enables $f_d$ to produce both highly structured and metric depth and still work in a fashion without groundtruth depth for training.  

We first estimate rough alignment factors of scale $a_s$ and shift $a_t$ from DPT's output $D^*_t$ to predicted depth $D_t$ by minimizing differences between $\bar{D}^*_t=a_s D^*_t + a_t$ and $D_t$ with closed-form expressions from the least-square optimization (see the supplementary).

\textbf{Statistical loss}. Compared with image-domain structures, depth-domain structures exclude depth-irrelevant low-level cues such as textures and painted patterns on objects and show geometric structures. Image structure similarity can be obtained by SSIM \cite{wang2002universal,wang2004image,wang2003multiscale} w.r.t. statistical constraints. Depth-domain structures also correlate to depth distribution represented by mean, variance, and co-variance for similarity measures. Thus, we compute the SSIM with depth map input $\bar{D}_t^*$ and $D_t$ and use the negative metric as the loss term 
\begin{equation}
     \mathcal{L}_{stat}= 1-\text{SSIM}(\bar{D}_t^*,D_t),
\label{loss_statistical}
\end{equation}
Unlike the widely-used appearance loss that combines SSIM with $L_1$ loss, we find that pixel-wise difference measures lead to unstable training since inverting from disparity to depth magnifies prediction uncertainty and produces much larger outliers in arbitrary ranges. In contrast, the SSIM loss constrains on the mean and variance terms for two distributions instead of per-pixel differences and becomes a desirable choice.

\begin{figure}[bt!]
    \centering
    \includegraphics[width=1.0\linewidth]{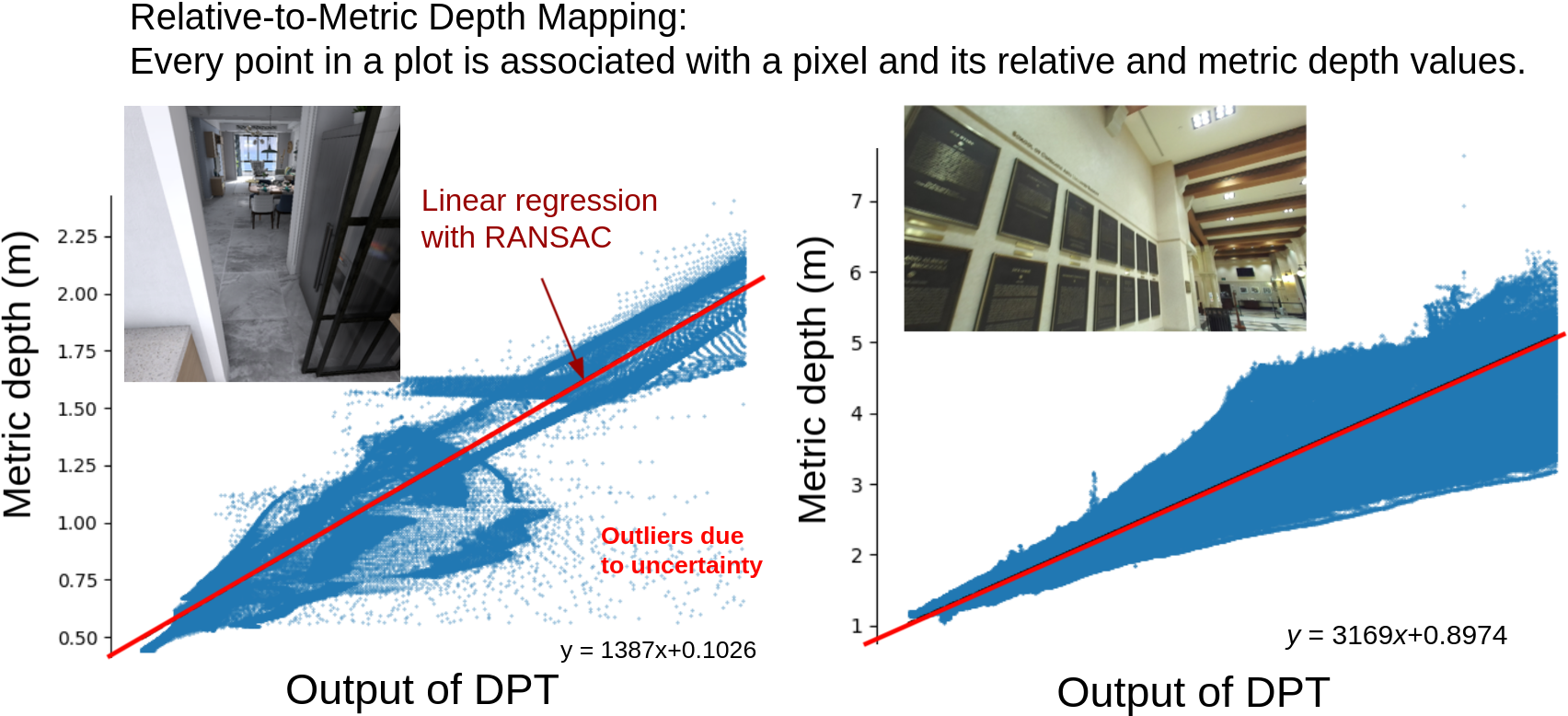}
    \vspace{-12pt}
    \caption{\textbf{Mapping from relative depth to metric depth is non-trivial.} $x$-axis captures (inverse) output values of DPT and $y$-axis represents metric depth from simulation systems or sensors. We run least-square linear regression with RANSAC to produce the optimal fitted lines between relative depth and metric depth for random scenes from our dataset.}
    \vspace{-8pt}
    \label{relation}
\end{figure}

\textbf{Spatial refinement loss}. SSIM loss only constrains statistical depth distribution but loses spatial information. We next propose a spatial control using depth occluding boundary maps (Fig. \ref{distdepth} (B)). The Sobel filter, which is a first-order gradient operator \cite{kanopoulos1988design}, is applied to compute depth-domain gradients: $g=(\frac{\partial X}{\partial u}, \frac{\partial X}{\partial v}),$ where $X \in \{\bar{D}_t^*, D_t\}$ and $u,v$ represent horizontal and vertical direction on 2D grids. Then we calculate a turn-on level $\alpha=\text{quantile}(\|g\|_2, 95\%)$ at the 95\%-quantile level of gradient maps to determine the depth occluding boundaries, where gradients are larger than $\alpha$. We compute the $0/1$ binary-value maps, $E^*$ and $E$, to represent occluding boundary locations by thresholding $\bar{D}_t^*$ and $D_t$ with their respective $\alpha$ terms. Last we calculate the Hamming distance, i.e. bitwise difference, of $E^*$ and $E$ and normalize it by the map size and use it as the spatial loss term,
\begin{equation}
    \vspace{-2pt}
     \mathcal{L}_{spat}= E^* \oplus E/|E|,
     \vspace{-2pt}
\label{loss_spatial}
\end{equation}
where $\oplus$ is the XOR operation for two boolean sets, and $|E|$ computes the size of a set. In implementation, to make thresholding and binarization operations differentiable, we substract respective $\alpha$ from $\bar{D}_t^*$ and $D_t$ and apply soft-sign function, which resembles the sign function but back-propagates smooth and non-zero gradients, to obtain maps with values in \{-1, 1\}. After the division by 2, we arrive at the element-wise Hamming distance between the maps. The loss function for structure distillation is $\mathcal{L}_{dist}=\mathcal{L}_{stat}+10^{-1}\mathcal{L}_{spat}$. The final loss function $\mathcal{L}_t$ for $I_t$ is combined with left-right  consistency $\mathcal{L}_{LR}= pe(I'_t, \hat{I}'_t)$, temporal consistency $\mathcal{L}_{temp}= pe(I_t, \hat{I}_{t+k \to t})$, where $\hat{I}_{t+k \to t}$ is forward warping and backward warping, $k \in \{1, -1\}$, and $\mathcal{L}_{dist}$:
\begin{equation}
\vspace{-3pt}
     \mathcal{L}_t = \mathcal{L}_{LR}+\mathcal{L}_{temp}+10^{-1}\mathcal{L}_{dist}.
\label{total_loss}
\vspace{-3pt}
\end{equation}

The designed structure distillation is key to gearing up the self-supervised depth estimator with high generalizability to unseen textures such that it better separates depth-relevant and depth-irrelevant low-level cues. From another perspective, the student trained by left-right consistency helps DPT learn ranges across different indoor scenes. 

Another alternative way is to predict scale and shift factors to align relative depth to metric depth based on the alignment relation \cite{Ranftl2020}. This seemly simple method, however, suffers from the disadvantage that depth estimation from neural networks inevitably includes uncertainty, which is either caused by neural network model or caused by data \cite{kendall2017uncertainties, kendall2018multi, maddox2019simple}. Conversion between relative depth and metric depth shows overall linear but noisy trends, and the optimal line equations can vary a lot for different scenes, as shown in Fig.~\ref{relation}. Thus, this alternative approach cannot factorize noise and outliers with only scale and shift terms. We show experiments using this approach in the supplementary.

We adopt ResNet \cite{he2016deep} as DepthNet $f_d$. Although one can use dense vision transformers for higher prediction accuracy, they suffer from low inference speed and cannot meet real-time on-device depth sensing due to larger network size and complex operations. Therefore, we maximally exploit structure knowledge embedded in DPT and also downsize the large vision transformer to smaller-size ResNet, which enables us to run depth sensing at an interactive rate (35$+$ fps v.s. 8-11 fps for different version DPT, measured on a laptop with RTX 2080 GPU) to fulfill the practical depth estimator purpose. See the supplementary for demonstration.

\begin{table*}[tb!]
\begin{center}
  \caption{\textbf{Quantitative comparison on the VA dataset.} Our DistDepth attains much lower errors than prior works of left-right consistency. DistDepth-M further uses the test-time multi-frame strategy in ManyDepth. See the main text.}
  %\scriptsize
  \footnotesize
  \label{table:UE4_comparison}
  \begin{tabular}[c]
  {
  p{1.0cm}<{\arraybackslash}|
  p{1.4cm}<{\centering\arraybackslash}|
  p{1.3cm}<{\centering\arraybackslash}|
  p{1.3cm}<{\centering\arraybackslash}|
  p{1.6cm}<{\centering\arraybackslash}|
  p{1.3cm}<{\centering\arraybackslash}|
  p{1.5cm}<{\centering\arraybackslash}|
  p{1.6cm}<{\centering\arraybackslash}}
  \hlineB{2}
  \hline
     & \multicolumn{4}{c}{\cellcolor[HTML]{FFFE65}Test-Time Single-Frame} & \multicolumn{3}{c}{\cellcolor[HTML]{67FD9A}Test-Time Multi-Frame} \\
      Method  & MonoDepth2 \cite{Godard_2019_ICCV} & DepthHints \cite{watson2019self} & DistDepth & Improvement & ManyDepth \cite{watson2021temporal} & DistDepth-M & Improvement\\
    \hline
      MAE    & 0.295 & 0.291 &\textbf{0.253} & -14.2\% & 0.275 & \textbf{0.239} & -13.1\% \\
      AbsRel & 0.203 & 0.197 &\textbf{0.175} & -13.8\% & 0.189 & \textbf{0.166} & -12.2\% \\
      RMSE   & 0.432 & 0.427 &\textbf{0.374} & -13.4\% & 0.408 & \textbf{0.357} & -12.5\% \\ 
      RMSE$_{log}$ & 0.251 & 0.248 & \textbf{0.213} & -15.1\% & 0.241 & \textbf{0.210} & -12.9\% \\
    \hlineB{2}
    \hline
  \end{tabular}
  \vspace{-20pt}
\end{center}
\end{table*}

\section{Datasets}
\label{sec:dataset}
\subsection{Training: SimSIN}
\label{sec:SimSIN}
To utilize the left-right and temporal neighboring frames to attain photometric consistency for self-supervised training, we adopt the popular Habitat simulator \cite{savva2019habitat, szot2021habitat} that initiates a virtual agent and renders camera-captured 3D indoor environments. We adopt Replica, MP3D, and HM3D as the backend 3D models following prior embodied AI works \cite{chen2020soundspaces, chen2021semantic, gao2020visualechoes, dean2020see, purushwalkam2020audio,chen2020learning}. 

We adopt a stereo baseline of 13cm following the camera setting in \cite{chabra2019stereodrnet} and render at 512 $\times$ 512 resolution. The agent navigates multiple times and captures stereo sequences. We then manually filter out failure sequences, such as when the agent gets too close to walls or navigates to null spaces. Our dataset consists about 80K, 205K, and 215K images from Replica, MP3D, and HM3D respectively, amounting to 500K stereo images from $\sim$1000 various environments in our proposed SimSIN dataset, which is by far the largest stereo dataset for generic indoor environments.

\subsection{Training: UniSIN}
\label{sec:UniSIN}
To investigate the gap between simulation and reality and compare performances of models trained on simulation and trained on real data. We use ZED-2i \cite{ZED2i}, a high-performing stereo camera system, to collect large-scale stereo sequences from various interior spaces around a university and create the UniSIN dataset. Its training split contains 500 sequences, and each sequence has 200 stereo pairs, amounting to 200K training images.

\subsection{Evaluation Sets}
\label{sec:evaluation}
\textbf{Commercial-Quality simulation}. We select a delicately designed virtual apartment (\textit{VA}) and render about 3.5K photorealistic images along a trajectory as the evaluation set \cite{UnrealEngine4, UE4Environment}. The VA dataset contains challenging indoor scenes for depth sensing, such as cabinet cubes with different lighting, thin structures, and complex decorators. These scenes enable us to conduct a detailed study of depth sensing in private indoor spaces, the most common use cases for AR/VR. We further include samples from pre-rendered \textit{Hypersim} \cite{roberts2020hypersim} dataset, which contains monocular images of virtual environments, for qualitative demonstration.

\textbf{Real Data}.
We adopt popular NYUv2 \cite{silberman2012indoor} whose test set contains 654 monocular images with depth maps from time-of-flight laser using Kinect v1. To compensate for Kinect's older imaging system and low resolution to serve more practical AR/VR use, we collect 1K high-definition images with finely optimized depth delivered by ZED-2i for numerical evaluation.

We show a sample collection for all the datasets in the supplementary.

\section{Experiments and Analysis}
\label{sec:analysis}

We set input size to 256$\times$256, batch size to 24, and epoch number to 10. Adam \cite{kingma2014adam} is used as the optimizer with an initial learning rate of 2$\times$10$^{-4}$ that drops by a factor of 10 at epoch 8 and 10. We adopt common data augmentation of color jittering and random flipping. We use ResNet50 for our PoseNet $f_p$ and ResNet152 for DepthNet $f_d$, and the same for MonDepth2, DepthHints, and ManyDepth for comparison in this section. DPT adopts large-size dense transformer network to apply on in-the-wild scenes. Thus we choose larger architecture for DepthNet to demonstrate generalizability but can still run in interactive frame rate (illustrated in the supplementary).

\subsection{Experiments on Simulation Data}
\label{sec:exp_sim}
We use SimSIN as the training dataset in Sec.~\ref{sec:exp_sim} and evaluate on various commercial-quality simulation data.

\textbf{Prior self-supervised methods trained on SimSIN}. We first directly train MonoDepth2, DepthHints, and ManyDepth on SimSIN following settings in their papers and show fitting on the training data and inference on VA in Fig.~\ref{prior_work} to investigate the generalizability. ManyDepth and DepthHints attain better results than MonoDepth2. Our DistDepth produces highly regularized structures with  robustness to unseen examples, w.r.t groundtruth. The range prediction also improves, which we believe is due to better structure occluding boundary reasoning.

\begin{figure}[bt!]
    \centering
    \includegraphics[width=0.95\linewidth]{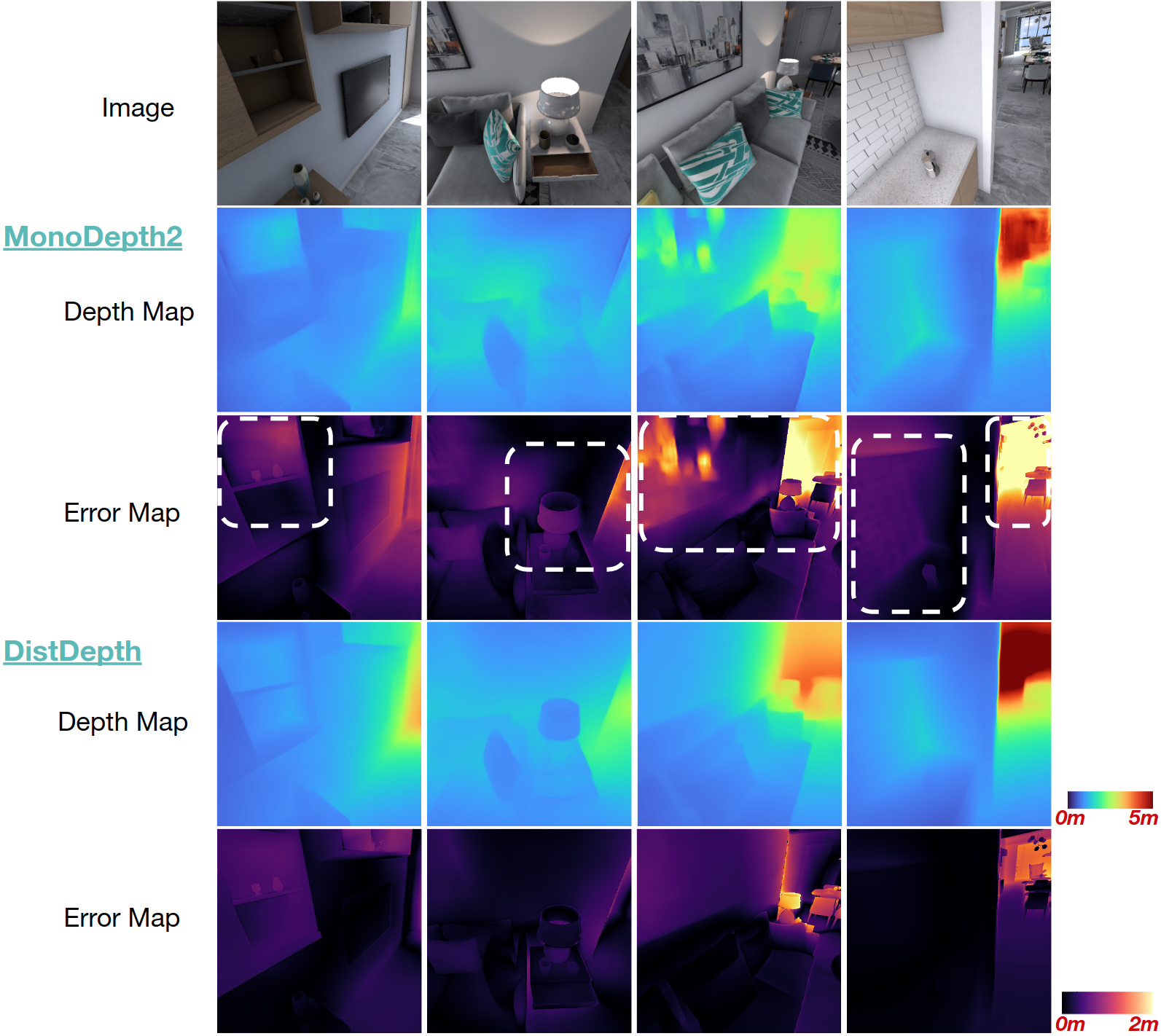}
    \vspace{-1pt}
    \caption{\textbf{Qualitative results on VA sequence.} Depth and error maps are shown for DistDepth and MonoDepth2 for comparison. These examples demonstrate that our DistDepth predicts geometrically structured depth for common indoor objects.}
    \vspace{-15pt}
    \label{results_ue4}
\end{figure}

\textbf{Error analysis on VA.}
The VA dataset includes various challenging scenes in indoor spaces. We show qualitative error analysis in Fig.~\ref{results_ue4}. Highlighted in error maps, our DistDepth has better generalizability on estimating the underlying geometric structures such as paintings, shelves, and walls under various lighting conditions. See supplementary for more examples.

We further show numerical comparison on the entire VA sequence in Table \ref{table:UE4_comparison}. 
All the methods in comparison are trained on SimSIN. We further equip DistDepth with the test-time multi-frame strategy with cost-volume minimization introduced in ManyDepth and denote this variant by DistDepth-M. 
Methods are categorized into test-time single-frame and test-time multi-frame. In both cases, DistDepth attains lower errors than prior arts. This validates our network design: with an expert used for depth-domain structure distillation, a student network $f_d$ can produce both structured and metric depth that is closer to the groundtruth.

\begin{figure}[bt!]
    \centering
    \includegraphics[width=0.92\linewidth]{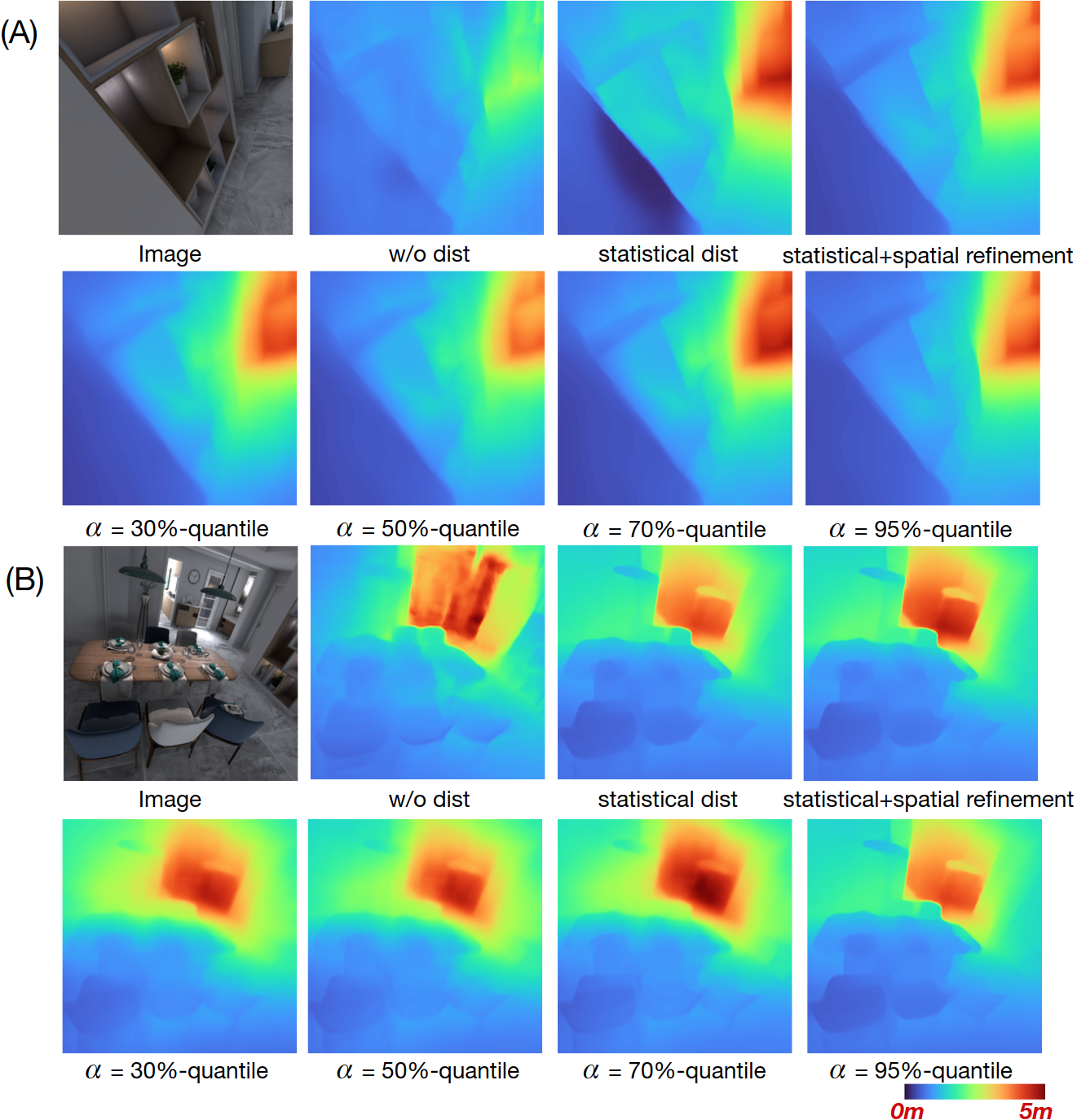}
    \vspace{-5pt}
    \caption{\textbf{Qualitative study for depth-domain structure improvement.} Two examples (A) and (B) are shown to study the effects of distillation (dist) losses and turn-on level $\alpha$ in spatial refinement to validate our design in Sec.~\ref{sec:distdepth}.}
    \vspace{-7pt}
    \label{ablation}
\end{figure}

\textbf{Ablation study on VA}. We first study the expert network and adopt different versions of DPT (hybrid and legacy), whose network sizes are different. Table~\ref{table:teacher_ablation} shows that the student network taught by the larger-size expert, DPT-legacy, achieves lower depth estimation errors. Without distillation, results are worse because its estimation relies only on the photometric loss, which fails on untextured areas like walls. As a sanity check, we also provide results of supervised training using SimSIN's groundtruth depth with pixel-wise MSE loss and test on the VA dataset, which shows the gap between training on curated depth and depth from expert network's predictions.

\begin{table}[tb!]
\begin{center}
  \caption{\textbf{Study on the choice of the expert network for distillation.} Different versions of DPT \cite{Ranftl2021} that vary in network sizes (\# of params) are adopted as the expert to teach the student. DPT-legacy localizes occluding contours better and leads to a better-performing student network. The results of supervised learning are provided as a reference.}
  \vspace{-8pt}
  \footnotesize
  \label{table:teacher_ablation}
  \begin{tabular}[c]
  {
  p{1.3cm}<{\arraybackslash}|
  p{1.2cm}<{\centering\arraybackslash}|
  p{1.2cm}<{\centering\arraybackslash}|
  p{1.2cm}<{\centering\arraybackslash}|
  p{1.3cm}<{\centering\arraybackslash}}
  \hlineB{2}
  \hline
  & \multicolumn{3}{c}{\cellcolor[HTML]{FFFE65}Self-Supervised} & \multicolumn{1}{c}{\cellcolor[HTML]{67FD9A}Supervised} \\
      Expert & w/o distillation & DPT - hybrid & DPT - legacy & with groundtruth \\
    \hline
      \# of params & - & 123M & 344M & -\\
      MAE & 0.295 & 0.276 & \textbf{0.253} & 0.221\\
      AbsRel & 0.203 & 0.188 & \textbf{0.175} & 0.158\\
      RMSE & 0.432 & 0.394 & \textbf{0.374} & 0.325\\ 
      RMSE$_{log}$ & 0.251 & 0.227 & \textbf{0.213} & 0.188\\
    \hlineB{2}
    \hline
  \end{tabular}
  \vspace{-25pt}
\end{center}
\end{table}

\begin{figure*}[bt!]
    \centering
    \includegraphics[width=0.83\linewidth]{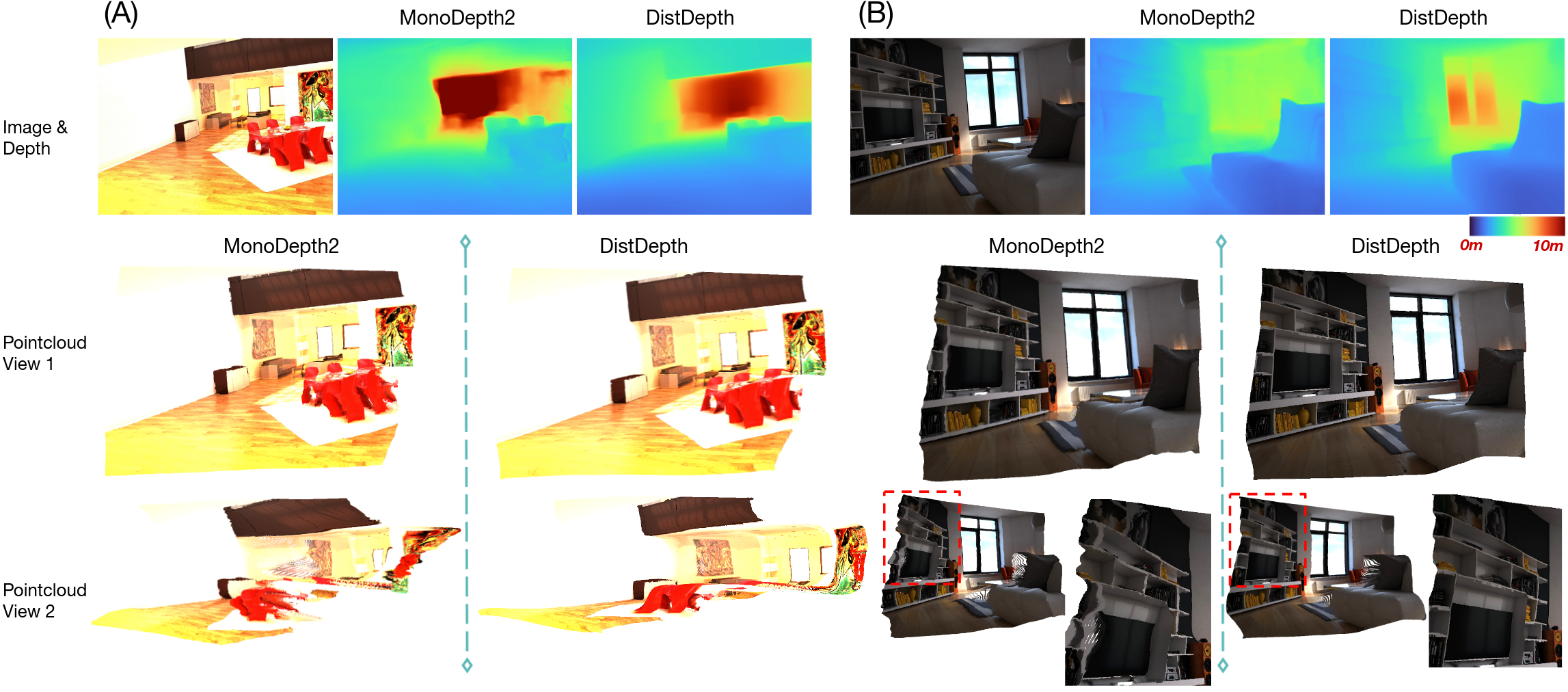}
    \vspace{-5pt}
    \caption{\textbf{Results on Hypersim.} Depth map and textured pointcloud comparison of MonoDepth2 and our DistDepth. With structure distillation, DistDepth attains better object structure predictions, such as tables and paintings on the wall shown in (A) and much less distortion for the large bookcase in (B).}
    \vspace{-10pt}
    \label{results_hypersim}
\end{figure*}

We next study the training strategy with different distillation losses and effects of turn-on level $\alpha$ in Sec.~\ref{sec:distdepth}. We compare (1) w/o distillation, (2) distillation with statistical loss only, and (3) distillation with statistical and spatial refinement loss. We demonstrate qualitative results in Fig.~\ref{ablation} to show the depth-domain structure improvements. Without distillation, spatial structures cannot be reasoned crisply. With statistical refinement, depth structures are more distinct. Adding spatial refinement, the depth-domain structures show fine-grained details. We further analyze the effects of different turn-on levels of $\alpha$. Low $\alpha$ makes structures blurry since the refinement does not focus on the high-gradient occluding boundaries as high $\alpha$ does, which identifies only high-gradient areas as occluding boundaries and benefits structure knowledge transfer.

\textbf{Comparison on Hypersim}. We next exhibit depth and textured pointcloud in Fig.~\ref{results_hypersim} for some scenes in Hypersim. Two different views are adopted for pointcloud visualization. One can find that our DistDepth predicts better geometric shapes in both depth map and pointcloud. See the supplementary for more examples.

\subsection{Experiments on Real Data}

\begin{figure*}[bt!]
    \centering
    \includegraphics[width=0.96\linewidth]{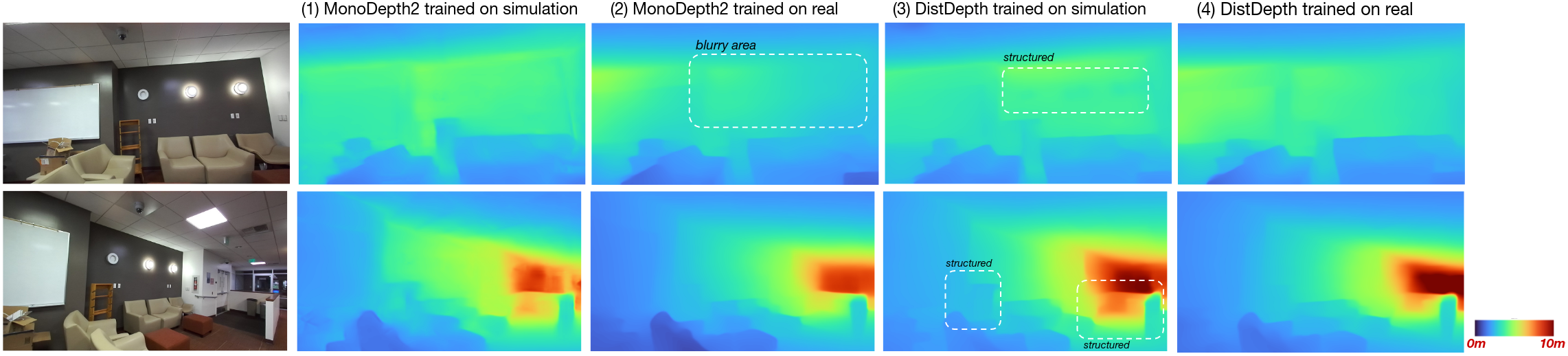}
    \vspace{-4pt}
    \caption{\textbf{Comparison on UniSIN.} Geometric shapes produced from DistDepth are better than MonoDepth2. DistDepth concretely reduces the gap for sim-to-real: (3) and (4) attain on-par results and sometimes training on simulation shows better structure than training on real.}
    \vspace{-5pt}
    \label{results_comparison}
\end{figure*}

\textbf{Closing sim-to-real gap}. We compare results of training on simulation (SimSIN)\footnote{To balance the training dataset size, we randomly subsample about 200K images here in SimSIN to match UniSIN's dataset size.} and real data (UniSIN) to investigate the performance gap.
We examine (1) training MonoDepth2 on simulation and evaluate on real data, (2) training MonoDepth2 on real data and evaluate on real data, (3) training DistDepth on simulation and evaluate on real data, and (4) training DistDepth on real and evaluate on real data. Fig.~\ref{results_comparison} illustrates the results of the four settings. Comparing (1) and (2), one can find that MonoDepth2 trained on real data produces more reliable results than on simulation. By contrast, this gap becomes unobvious when comparing (3) and (4) using DistDepth. Results of (3) are on-par with (4) and sometimes even produce better geometric shapes like highlighted areas. We further include numerical analysis in the supplementary.

\begin{figure*}[bt!]
    \centering
    \includegraphics[width=0.96\linewidth]{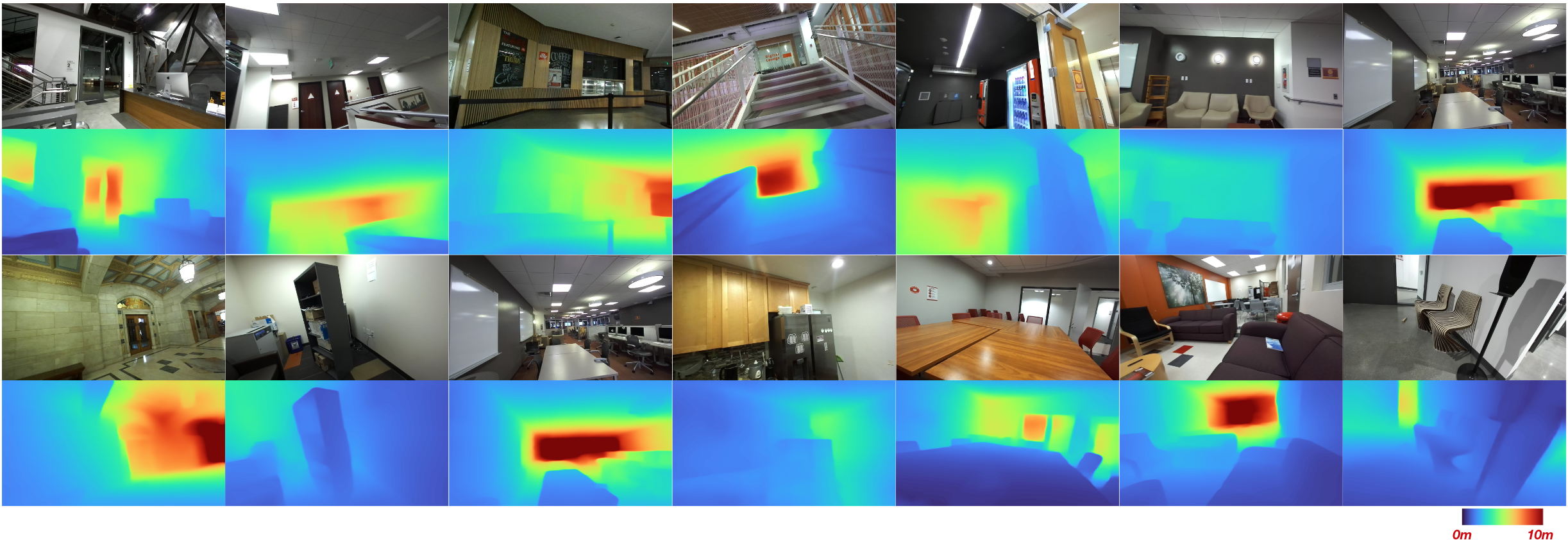}
    \vspace{-15pt}
    \caption{\textbf{Results on real data (UniSIN) using our DistDepth only trained on simulation (SimSIN).}}
    \vspace{-5pt}
    \label{results_university}
\end{figure*}

The results validate our proposals on both method and dataset levels. First, DistDepth utilizes an expert network to distill knacks to the student. The distillation substantially adds robustness to models trained on simulation data and makes the results comparable to models trained on real data. This shows the ability of DistDepth for \textit{closing the gap between simulation and real data}. Second, stereo simulation data provide a platform for left-right consistency to learn metric depth from stereo triangulation. We show a collection of results in Fig.~\ref{results_university} using DistDepth that is trained purely on simulation.

\textbf{Evaluation on NYUv2}. Table~\ref{table:nyuv2} shows evaluations on NYUv2.
We first train our DistDepth on SimSIN and finetune on NYUv2 with only temporal consistency.
Note that one finetuned model (Sup:$\triangle$) is categorized as semi-supervised since it utilizes an expert that has been trained with NYUv2's curated depth. The finetuned models produce the best results among  methods without NYUv2's depth supervision and even attain comparable results to many supervised methods.
We next train DistDepth only on simulation (SimSIN) or real data (UniSIN) and evaluate on NYUv2.
Performances of the model trained on SimSIN only drop a little compared with that trained on UniSIN, which justifies our sim-to-real advantage again.
Without involving any training data in NYUv2, DistDepth still achieves comparable performances to many supervised and self-supervised methods, which further validates our \textit{zero-shot cross-dataset advantage}. 
We exhibit real-time depth sensing, 3D photos, and depth-aware AR applications in supplementary. 

\begin{table}[tb!]
\begin{center}
  \caption{\textbf{Evaluation on NYUv2.} Sup: \cmark- supervised learning using groundtruth depth, \xmark- not using groundtruth depth, and $\triangle$- semi-supervised learning (we use the expert finetuned on NYUv2, where we have indirect access to the groundtruth). We achieve the best results among all self-supervised methods, and our semi-supervised and self-supervised finetuned on NYUv2 even outperform many supervised methods. The last two rows show results without groundtruth supervision and without training on NYUv2. In this challenging \textit{zero-shot cross-dataset evaluation}, we still achieve comparable performances to many methods trained on NYUv2. Error and accuracy (yellow/green) metrics are reported.}
  \vspace{-5pt}
  \scriptsize
  \label{table:nyuv2}
  \begin{tabular}[c]
  {
  p{2.14cm}<{\arraybackslash}|
  p{0.25cm}<{\centering\arraybackslash}|
  p{0.86cm}<{\centering\arraybackslash}|
  p{0.59cm}<{\centering\arraybackslash}|
  p{0.5cm}<{\centering\arraybackslash}|
  p{0.29cm}<{\centering\arraybackslash}|
  p{0.29cm}<{\centering\arraybackslash}|
  p{0.29cm}<{\centering\arraybackslash}}
  \hlineB{2}
  \hline
      Methods & Sup & Train on NYUv2 & \cellcolor[wave]{580} AbsRel& \cellcolor[wave]{580} RMSE& \cellcolor[wave]{500}$\delta_1$ & \cellcolor[wave]{500}$\delta_2$ & \cellcolor[wave]{500}$\delta_3$\\
    \hline
      Make3D \cite{saxena2008make3d} & \cmark & \cmark & 0.349 & 1.214 & 44.7 & 74.5 & 89.7 \\  
      Li $\textit{et al.}$ \cite{li2017two} & \cmark & \cmark & 0.143 & 0.635 & 78.8 & 95.8 & 99.1 \\
      Eigen $\textit{et al.}$ \cite{eigen2015predicting} & \cmark & \cmark & 0.158 & 0.641 & 76.9 & 95.0 & 98.8\\
      Laina $\textit{et al.}$ \cite{laina2016deeper} & \cmark & \cmark & 0.127 & 0.573 & 81.1 & 95.3 & 98.8 \\
      DORN \cite{fu2018deep} & \cmark & \cmark & 0.115 & 0.509 & 82.8 & 86.5 & 99.2 \\
      AdaBins \cite{bhat2021adabins} & \cmark & \cmark & \textbf{0.103} & 0.364 & 90.3 & 98.4 & 99.7\\ 
      DPT \cite{Ranftl2021} & \cmark & \cmark & 0.110 & \textbf{0.357} & \textbf{90.4} & \textbf{98.8} & \textbf{99.8}\\ 
      \hline
      Zhou $\textit{et al.}$ \cite{zhou2019moving} & \xmark & \cmark & 0.208 & 0.712 & 67.4 & 90.0 & 96.8 \\  
      Zhao $\textit{et al.}$ \cite{zhao2020towards}  & \xmark & \cmark & 0.189 & 0.686 & 70.1 & 91.2 & 97.8 \\
      Bian $\textit{et al.}$ \cite{bian2021unsupervised} & \xmark & \cmark & 0.157 & 0.593 & 78.0 & 94.0 & 98.4\\
      P$^2$Net+PP \cite{yu2020p} & \xmark & \cmark & 0.147 & 0.553 & 80.4 & 95.2 & 98.7 \\
      StructDepth \cite{li2021structdepth} & \xmark & \cmark & 0.142 & 0.540 & 81.3 & 95.4 & 98.8 \\
      MonoIndoor \cite{ji2021monoindoor} & \xmark & \cmark & 0.134 & 0.526 & 82.3 & 95.8 & 98.9 \\
      DistDepth (finetuned) & 	\xmark & \cmark & \textbf{0.130} & \textbf{0.517} & \textbf{83.2} & \textbf{96.3} & \textbf{99.0}\\
      DistDepth (finetuned) & 	$\triangle$ & \cmark & 0.113 & 0.444 & 87.3 & 97.4 & 99.3\\
      \hline
      DistDepth (SimSIN) & \xmark & \xmark & 0.164 & 0.566 & 77.9 & 93.5 & 98.0\\
      DistDepth (UniSIN) & \xmark & \xmark & \textbf{0.158} & \textbf{0.548} & \textbf{79.1} & \textbf{94.2} & \textbf{98.5}\\
    \hlineB{2}
    \hline
  \end{tabular}
  \vspace{-21pt}
\end{center}
\end{table}

\section{Conclusion and Discussion}
This work targets at a practical indoor depth estimation framework with following features: training without depth groundtruth, effective training on simulation, high generalizability, and accurate and real-time inference.  
We first identify the challenges of indoor depth estimation and study the applicability of existing self-supervised methods with left-right consistency on SimSIN. Geared up with the depth-domain structure knowledge distilled from an expert, we see substantial improvement in both inferring finer structures and more accurate metric depth. 
We show zero-shot cross-dataset inference that proves its generalizability to work on heterogeneous data domains and attain a broadly applicable depth estimator for indoor scenes. Even more, depth learned from simulation data transfers well to real scenes, which shows the success of our distillation strategy. At inference time, it only takes a single feed-forward pass to DepthNet to produce structured metric depth and reach 35$+$ fps on a portable device which serves real-time needs. 

\textbf{Limitations}. Although DistDepth is capable of producing structured and metric depth using a single forward pass of depth estimation, it operates on a per-frame basis, which can be refined to produce more temporally consistent depth for video inputs \cite{luo2020consistent, Kopf_2021_CVPR}. Another issue commonly for depth estimation is the proper handling of reflective objects. With distillation, DistDepth can produce estimation for objects with clear contours, as illustrated in Fig.~\ref{teaser} of bulbs. However, our approach is still not yet robust to large mirrors. A possible solution is to locate mirrors and perform depth completion \cite{zhong2019deep, zhu2021rgb, wu2021scene, wong2021unsupervised} on raw estimates.

{\small
\bibliographystyle{ieee_fullname}
\bibliography{egbib}
}

\end{document}